\documentclass{article}

\PassOptionsToPackage{numbers, compress}{natbib}

\usepackage[preprint]{neurips_2022}

\usepackage[utf8]{inputenc} 
\usepackage[T1]{fontenc}    
\usepackage{hyperref}       
\usepackage{url}            
\usepackage{booktabs}       
\usepackage{amsfonts}       
\usepackage{nicefrac}       
\usepackage{microtype}      
\usepackage{xcolor}         
\usepackage{comment}
\usepackage{graphicx}
\usepackage{amsmath}
\usepackage{bm}
\usepackage{algorithm}
\usepackage{algpseudocode}
\usepackage{multirow}
\usepackage{multicol}

\usepackage{subcaption}
\usepackage{colortbl}
\usepackage{amsthm}
\usepackage{thmtools,thm-restate}

\DeclareMathOperator*{\argmax}{argmax}

\newlength\savewidth\newcommand\shline{\noalign{\global\savewidth\arrayrulewidth
  \global\arrayrulewidth 1pt}\hline\noalign{\global\arrayrulewidth\savewidth}}

\usepackage{amsmath}
\DeclareMathOperator*{\argmin}{argmin} 

\usepackage{xspace}

\makeatletter
\DeclareRobustCommand\onedot{\futurelet\@let@token\@onedot}
\def\@onedot{\ifx\@let@token.\else.\null\fi\xspace}

\def\eg{\emph{e.g}\onedot}

\def\etal{\emph{et al}\onedot}
\makeatother

\title{Improved Vector Quantized Diffusion Models}

\author{%
  Zhicong Tang\textsuperscript{\textnormal{1}}\qquad
  Shuyang Gu\textsuperscript{\textnormal{2}}\qquad
  Jianmin Bao\textsuperscript{\textnormal{3}}\qquad\\
  \textbf{Dong Chen}\textsuperscript{\textnormal{3}}\qquad
  \textbf{Fang Wen}\textsuperscript{\textnormal{3}}\qquad\\ \\
  \textsuperscript{1}Tsinghua University \\
  \textsuperscript{2}University of Science and Technology of China \\
  \textsuperscript{3}Microsoft Research \\ \\
  \texttt{tzc21@mails.tsinghua.edu.cn} \\
  \texttt{gsy777@mail.ustc.edu.cn} \\
  \texttt{\{jianbao,doch,fangwen\}@microsoft.com}
}

\begin{document}

\maketitle

\begin{abstract}

Vector quantized diffusion (VQ-Diffusion) is a powerful generative model for text-to-image synthesis, but sometimes can still generate low-quality samples or weakly correlated images with text input. We find these issues are mainly due to the flawed sampling strategy. In this paper, we propose two important techniques to further improve the sample quality of VQ-Diffusion. 1) We explore classifier-free guidance sampling for discrete denoising diffusion model and propose a more general and effective implementation of classifier-free guidance. 2) We present a high-quality inference strategy to alleviate the joint distribution issue in VQ-Diffusion.  
Finally, we conduct experiments on various datasets to validate their effectiveness and show that the improved VQ-Diffusion suppresses the vanilla version by large margins. We achieve an \textbf{8.44} FID score on MSCOCO, surpassing VQ-Diffusion by \textbf{5.42} FID score. When trained on ImageNet, we dramatically improve the FID score from \textbf{11.89} to \textbf{4.83}, demonstrating the superiority of our proposed techniques. The code is released at \href{https://github.com/microsoft/VQ-Diffusion}{\texttt{https://github.com/microsoft/VQ-Diffusion}}.

\end{abstract}

\section{Introduction}
Denoising Diffusion Probability Models(DDPM)~\cite{ho2020denoising} have shown remarkable success in various fields, such as, image generation~\cite{gu2021vector,nichol2021glide,ramesh2022hierarchical}, video generation~\cite{ho2022video}, audio generation~\cite{kong2020diffwave} and so on. Depending on the signal type, DDPM could be roughly divided into continuous diffusion models~\cite{ho2020denoising,nichol2021improved} and discrete diffusion models~\cite{austin2021structured,gu2021vector}. In the forward step, the former adds Gaussian noise on continuous signals, while the latter uses a Markov Transition matrix to obfuscate discrete input tokens.

Extensive prior studies have focused on improving the DDPM from various aspects, including better network architecture~\cite{nichol2021glide}, hierarchical structure design~\cite{ramesh2022hierarchical}, alternative loss function~\cite{huang2021variational}, fast sampling strategy~\cite{kong2021fast}, \etal. The current state-of-the-art models are already capable of producing photo-realistic images, like GLIDE~\cite{nichol2021glide} and DALL-E 2~\cite{ramesh2022hierarchical} have presented impressive results in text-to-image synthesis. However, most previous improvements are based on continuous diffusion models and very few attempts are on discrete discrete diffusion models.

In this paper, we aim to improve the sample quality of discrete diffusion models, more specifically VQ-Diffusion~\cite{gu2021vector}, which leverages the VQVAE~\cite{oord2017neural} to encode images to discrete tokens and then perform diffusion process in discrete space. One of the major strengths of VQ-Diffusion is that we can estimate the probability for each discrete token, thus it achieves high-quality images with relatively fewer inference steps.
Based on this, we introduce several techniques intended to improve VQ-Diffusion.

\noindent \textbf{Discrete classifier-free guidance}. 
For conditional image generation, suppose the condition information is $y$, and the generated image is $x$. The diffusion generative models try to maximize prior probability $p(x|y)$, and assume the generated images $x$ will satisfy the constraints of posterior probability $p(y|x)$. However, we found this assumption may fail and ignore the posterior probability in most cases. We name this as the \emph{posterior issue}. To address this issue, we propose to take both the prior and posterior into consideration simultaneously. Powered by posterior constraint, the generated images are significantly improved in terms of quality and consistency with input conditions. This approach shares the spirit of the previous classifier-free technique~\cite{ho2021classifier}. However, our methods are formulated more precisely since our model estimates probability instead of noise. Besides, instead of setting input conditions to zero, we introduce a more general and effective implementation of classifier-free guidance by using a learnable parameter as a condition to approximate $p(x)$. We found it could further improve the performance.

\noindent \textbf{High-quality inference strategy}. In each denoising step, we usually sample multiple tokens simultaneously and each token is sampled with estimated probability independently. However, different locations are often associated, thus sampling independently may ignore the dependencies. Assuming a simple dataset with only two samples: AA and BB. Each sample has $50\%$ chances to appear. However, if we sample independently based on the estimated probability of each location, incorrect outputs (AB and BA) will appear, even if these samples never appear during training. We call this the \emph{joint distribution issue}. To alleviate this issue, we introduce a high-quality inference strategy. It is based on two core designs. First, we reduce the number of sampled tokens at each step since more sampled tokens would suffer from the joint distribution issue more heavily. Second, we find that tokens with high confidence tend to be more accurate, thus we introduce purity prior to sample tokens with high confidence.

Powered by these techniques, we improve the sampling quality of VQ-Diffusion by large margins. We conduct experiments on CUB-200, MSCOCO, Conceptual Captions, and an even larger Internet dataset, and find that by fixing the posterior issue and the joint distribution issue, VQ-Diffusion could notably improve its performance. Concretely, we achieve an \textbf{8.44} FID score on MSCOCO, surpassing VQ-Diffusion by \textbf{5.42} FID score. When trained on class conditional ImageNet dataset, we dramatically improve the FID score from \textbf{11.89} to \textbf{4.83}. Above all, our key contribution contains three parts:
\begin{enumerate}
    \item We find adding the posterior constraint will improve the quality of generated images significantly, and introduce a more general and effective implementation of classifier-free guidance.

    \item We point out the joint distribution issue in VQ-Diffusion, and propose the High-quality sampling strategy to alleviate it.
    
    \item We validate our approaches on various datasets, demonstrate these techniques improve VQ-Diffusion to achieve the state-of-the-art performance on various tasks.
\end{enumerate}

\section{Background: VQ-Diffusion}
We first briefly review vector quantized diffusion (VQ-Diffusion) models and analyze the reason for the existence of joint distribution issue. VQ-Diffusion starts with a VQVAE that converts images $x$ to discrete tokens $x_0 \in \{1,2,...,K, K+1\}$, $K$ is the size of codebook, and $K+1$ denotes the $[\text{\tt{MASK}}]$ token. Then the forward process of a diffusion model $q(x_t|x_{t-1})$ is a Markov chain that adds noise at each step. The reverse denoising process recovers the sample from a noise state. Specifically, the forward process is given by:
\begin{equation}
q(x_t|x_{t-1}) = \bm{v}^\top(x_{t})\bm{Q}_t \bm{v}(x_{t-1})
\label{eqn:markov}
\end{equation}
where $\bm{v}(x)$ is a one-hot column vector with entry 1 at index $x$. And $\bm{Q}_t$ is the probability transition matrix from $x_{t-1}$ to $x_t$. Specifically, for the mask-and-replace VQ-diffusion strategy,
\begin{equation}
   \bm{Q}_t =
   \begin{bmatrix}
   \alpha_t + \beta_t & \beta_t & \beta_t & \cdots & 0 \\
   \beta_t & \alpha_t + \beta_t & \beta_t & \cdots & 0 \\
   \beta_t & \beta_t & \alpha_t + \beta_t & \cdots & 0 \\
   \vdots & \vdots & \vdots & \ddots & \vdots \\
   \gamma_t & \gamma_t &\gamma_t  & \cdots & 1 \\
   \end{bmatrix}.
\label{eqn:mask_transit}
\end{equation}

given $\alpha_t \in [0,1]$, $\beta_t=(1-\alpha_t-\gamma_t)/K$ and $\gamma_t$ the probability of a token to be replaced with a $[\text{\tt{MASK}}]$ token.

The reverse process is given by the posterior distribution:
\begin{equation}
    q(x_{t-1}|x_t,x_0) = \frac{(v^{T}(x_t)\bm{Q}_{t}v(x_{t-1}))(v^{T}(x_{t-1})\bar{\bm{Q}}_{t-1}v(x_{0}))}{v^{T}(x_t)\bar{\bm{Q}}_{t}v(x_{0})} 
    \label{eqn:posterior}
\end{equation}
where $\bar{\bm{Q}}_t=\bm{Q}_t\cdots\bm{Q}_1$. The cumulative transition matrix $\bar{\bm{Q}}_t$ and the probability $q(x_t|x_0)$ can be computed in closed form with:
\begin{equation}
	\bar{\bm{Q}}_t \bm{v}(x_0)=\bar{\alpha}_t \bm{v}(x_0) + (\bar{\gamma}_t - \bar{\beta}_t)\bm{v}(K+1) + \bar{\beta}_t
	\label{eqn:fast_Qtv}
\end{equation}
Where $\bar{\alpha}_t=\prod_{i=1}^t\alpha_i$, $\bar{\gamma}_t=1-\prod_{i=1}^t(1-\gamma_i)$, and $\bar{\beta}_t = (1-\bar{\alpha}_t-\bar{\gamma}_t)/(K+1)$ can be calculated and stored in advance.

In the inference time, the denoising network $p_\theta$ gradually recovers the corrupted input via a fixed Markov chain. 
Besides, to stable the training and enable a fast inference strategy, VQ-Diffusion proposes a reparameterization trick that the denoising network predicts the denoised token distribution $p_\theta(\tilde{\bm{x}}_0 | \bm{x}_t)$ at each step. Thus we can compute the reverse transition distribution according to:
\begin{equation}
	\!\!\!\! p_\theta(\bm{x}_{t-1}|\bm{x}_t) = \sum_{\tilde{\bm{x}}_0=1}^{K} q(\bm{x}_{t-1}|\bm{x}_t,\tilde{\bm{x}}_0) p_\theta(\tilde{\bm{x}}_0|\bm{x}_t).
	\label{eqn:Reparameter_trick}
\end{equation}

We observed that VQ-Diffusion may suffer from the following two issues: 
1) For conditional image generation, \eg, text-to-image generation, the condition information $y$ directly injected into the denoising network $p_\theta(\bm{x}_{t-1}|\bm{x}_t, y)$, and then the network is hoped to use both $\bm{x}_t$ and $y$ to recover $x_{x-1}$. However, the network may ignore $y$, since $x_t$ already contains sufficient information. Thus the generated image may not correlate to the input $y$ well, causing the posterior issue.
2) For $t$-th timestep, each location of $\bm{x_{t-1}}$ is sampled from $p_\theta(\bm{x}_{t-1}|\bm{x}_t)$ independently. Thus, it could not model the correspondence among different locations. So sampling from this distribution parallel may be unreasonable, causing the joint distribution issue.

In the next section, we discuss these issues and propose a general classifier-free sampling strategy in the training procedure to solve the posterior constraint issue. Besides, we propose the high-quality inference strategy to alleviating the joint distribution issue in the inference stage.

\section{Method}
For the mismatch issue between generated image and input text, we propose the discrete classifier-free guidance as a posterior constraint to solve it. We also find a predefined prior on discrete space can help the sampling process and alleviate the joint distribution issue. Furthermore, we introduce the high-quality inference strategy to improve the sampling quality. We describe these techniques in the following section.

\subsection{Discrete Classifier-free Guidance}
For conditional image generation tasks like text-to-image synthesis, a mandatory requirement is that the generated images should match the condition input. VQ-Diffusion simply injects the condition information into the denoising network and suppose that the network will use both corrupted input and text to recover the original image. However, since the corrupted input usually contains much more information than the text, the network may ignore the text in the training phase. Thus, we find the VQ-Diffusion may easily generate images with poor correlation with input text, which is often calculated with CLIP score~\cite{radford2021learning}.

From the perspective of the optimization target, the diffusion model aims to find $x$ to maximize $p(x|y)$. However, a higher CLIP score also needs $p(y|x)$ as large as possible. Thus, a straight forward solution is to optimize $\log p(x|y) + s \log p(y|x)$, where $s$ is a hyper-parameter to control the degree of posterior constraint. Using Bayes' theorem, we can derive this optimization target as follows:
\begin{equation}
\begin{split}
&\argmax_x [\log p(x|y) + s \log p(y|x)]\\
=&\argmax_x [(s+1)\log p(x|y) - s \log \frac{p(x|y)}{p(y|x)}]  \\
=&\argmax_x [(s+1)\log p(x|y) - s \log \frac{p(x|y)p(y)}{p(y|x)}] \\
=&\argmax_x [(s+1)\log p(x|y) - s \log p(x)] \\ 
=&\argmax_x [\log p(x) + (s+1)(\log p(x|y) - \log p(x))]  \\ 
\end{split}
\label{eqn:cfs}
\end{equation}

To predict the unconditional image logits $p(x)$, a direct way is to fine-tune the model with a certain percentage of empty condition inputs, like GLIDE~\cite{nichol2021glide} which set the input condition to a ``null'' text to fine-tune the model. However, we find using a learnable vector instead of text embedding of ``null'' can better fit the logits of $p(x)$. In the inference stage, we first generate the conditional image logits $p_\theta(x_{t-1} | x_t, y)$, then predict the unconditional image logits $p_\theta(x_{t-1} | x_t)$ by setting conditional input to the learnable vector. The next denoising step samples from:

\begin{equation}
    \log p_\theta(x_{t-1} | x_t, y) = \log p_\theta(x_{t-1} | {x_t}) + (s+1) (\log p_\theta(x_{t-1} | x_t, y) - \log p_\theta( x_{t-1} | x_t)),
\label{eqn:cfsi}
\end{equation}

Compared with previous classifier-free sampling in continuous domain, our posterior constraint at discrete domain has three main difference: (1) First, since VQ-Diffusion leverage the reparameterization trick to predict $p(x|y)$ at unoised state, we may also apply Equation~\ref{eqn:cfs} on unnoised state, so it's compatible with other techniques like fast inference strategy~\cite{gu2021vector} or high-quality inference strategy (Sec.~\ref{sec:SIS}). (2) Second, diffusion models at continuous setting do not predict the probability $p(x|y)$ directly; they use gradient to approximate it. However, discrete diffusion models estimate the probability distribution directly. (3) Third, continuous models set condition to null vector to predict $p(x)$. We find using a learnable vector instead of null vector could further improve the performance. 

\subsection{High-quality Inference Strategy}
\label{sec:SIS}
Another important issue in VQ-Diffusion is the joint distribution issue that caused by sampling token of different locations independently. This may ignore the correlation between different positions. To alleviate this issue, we propose a high-quality inference strategy, which includes two key techniques. 

\noindent \textbf{Fewer tokens sampling.}
First, we propose to sample fewer tokens at each step. In this way, we model the correlation between different positions by the iterative denoising process rather than ignoring it when sampling multiple tokens independently. Concretely, the number of changed tokens in each step of VQ-Diffusion is uncertain. For simplicity, we set the changed tokens to a certain number of tokens in each step.

\begin{figure*}[t]
\centering
\includegraphics[width=1.0\columnwidth]{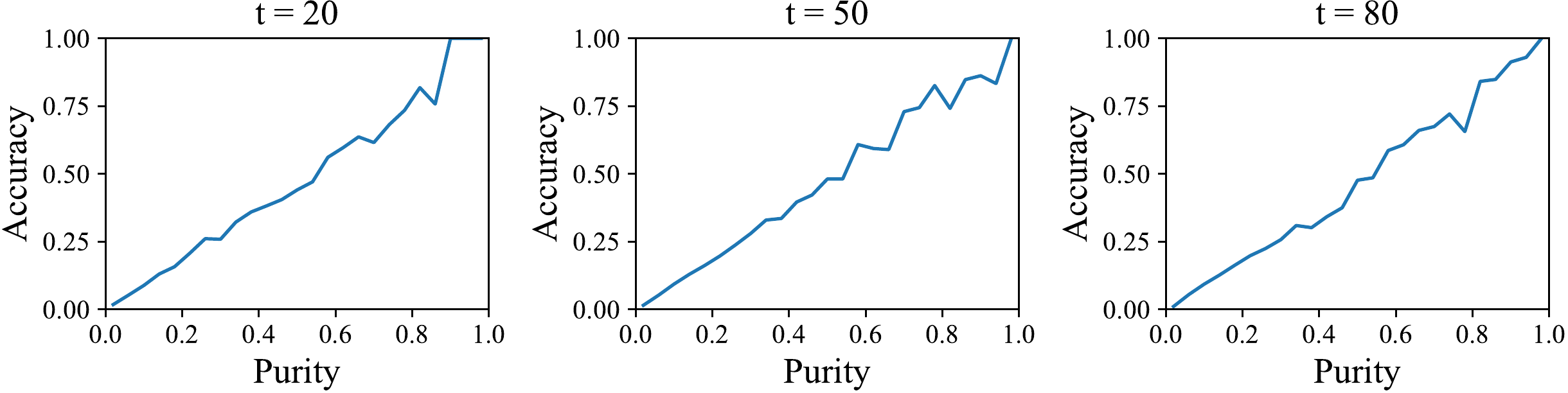}
\vspace{-0.5cm}
\caption{Illustration of the correlation between purity and accuracy of tokens at different timesteps(t=20, 50, and 80). We find high purity usually yields high accuracy.}
\label{fig:prior}
\end{figure*}

For the state of each step, we can count its number of mask and choose the proper timestep as the timestep embedding. Suppose the input is $x_t$, we have two sets: $A_t:=\{i|x_t^i = [\text{\tt{MASK}}]\}$, and $B_t:=\{i|x_t^i \neq [\text{\tt{MASK}}]\}$. We aim to recover $\Delta_z$ $[\text{\tt{MASK}}]$ tokens from $A_t$ in each step.  Thus, the total inference steps are $T' = (H \times W)/\Delta_z$, $H$ and $W$ indicate the spatial resolution of tokens. Current timestep $t$ could be calculated with $\argmin_{t}||\frac{|A_t|}{H\times W} - \bar{\gamma_t}||_2$, where $|A|$ denotes number of elements in set $A$. When $\Delta_z = 1$, it has the same inference speed as autoregressive models. Our fewer tokens sampling is the opposite of previous broadly studied fast sampling strategy~\cite{gu2021vector}. We seek to achieve high sampling quality by sacrificing inference time.

\noindent \textbf{Purity prior sampling.}
From Equation~\ref{eqn:posterior}, we can derive the following lemma:
\begin{restatable}{lemma}{lemmamask}
For any position $i$ which satisfied $x_t^i = [\text{\tt{MASK}}]$, then $q(x_{t-1}^i=[\text{\tt{MASK}}] | x_t^i = [\text{\tt{MASK}}], x_0^i)=  \bar{\gamma}_{t-1}/\bar{\gamma}_t$ is a constant.
\label{lemma:mask}
\end{restatable}
We leave the proof in the supplementary material. This lemma demonstrates that each position has the same probability to leave the $[\text{\tt{MASK}}]$ state. In other words, the transformation from $[\text{\tt{MASK}}]$ state to non-$[\text{\tt{MASK}}]$ state is position independent. However, we find different positions may have different confidence to leave $[\text{\tt{MASK}}]$ state. Specifically, positions with higher purity usually have higher confidence. We present the correlation between purity and accuracy in Figure~\ref{fig:prior}. We find that a higher purity score usually indicates a more accurate token. Therefore, our key idea is conducting importance sampling rely on the purity score instead of random sampling. By leveraging this purity prior, each step could sample tokens from a more confident region, thus improving the sampling quality. The definition of purity at location $i$ and timestep $t$ is:
\begin{equation}
	purity(i, t) = \max_{j=1...K} p(x_0^i=j|x_t^i)
	\label{eqn:purity}
\end{equation}

Based on these two techniques, we enable a high-quality inference strategy. 

\section{Experiments}
\subsection{Implementation details}
\paragraph{Datasets} To demonstrate the capability of our proposed techniques, we conduct experiments on three commonly used text-to-image synthesis datasets: CUB-200~\cite{wah2011caltech}, MSCOCO~\cite{lin2014microsoft}, and Conceptual Captions (CC)~\cite{sharma2018conceptual,changpinyo2021conceptual}. Instead of using all the data from Conceptual Captions dataset, we follow the setting in ~\cite{gu2021vector}, using a more balanced subset that contains 7M text-image pairs. Besides, to further demonstrate the scalability of our method, we collect 200 million high-quality text-images pairs from the internet. We named it ITHQ-200M dataset.

\paragraph{Backbone} For a fair comparison with the original VQ-Diffusion and other previous text-to-image methods under the similar number of parameters, we build two different backbone settings: 
1) Improved VQ-Diffusion-B (base), which consists of 370M parameters. We follow the network structure of VQ-Diffusion and directly use its released model as the pretrained model and fine-tune on each database. 
2) Improved VQ-Diffusion-L (large), which consists of 1.27B parameters. The image decoder contains 36 transformer blocks with dimensions of 1408. To achieve a more general text-to-image generation model, we train this large model on the ITHQ-200M dataset. We adopt the base size model on other datasets for most experiments.

\paragraph{Evaluation metrics} We use four metrics to evaluate the generated images. 1) FID score, which evaluates both quality and diversity of generated images. 2) Clip score~\cite{radford2021learning}, which measures the similarity between the generated image and text. 3) Quality Score(QS)~\cite{gu2020giqa}, which only measures the image quality. A higher QS score denotes higher image quality. 4) Diversity Score(DS), defined as $1-\text{DDS}$ where DDS denotes Diversity Difference Score proposed in ~\cite{gu2020priorgan}. It measures the diversity of generated images and a higher DS denotes more a diverse generated distribution.  

\begin{figure}[t]
\centering
\begin{minipage}[c]{0.495\linewidth}
\includegraphics[width=\linewidth]{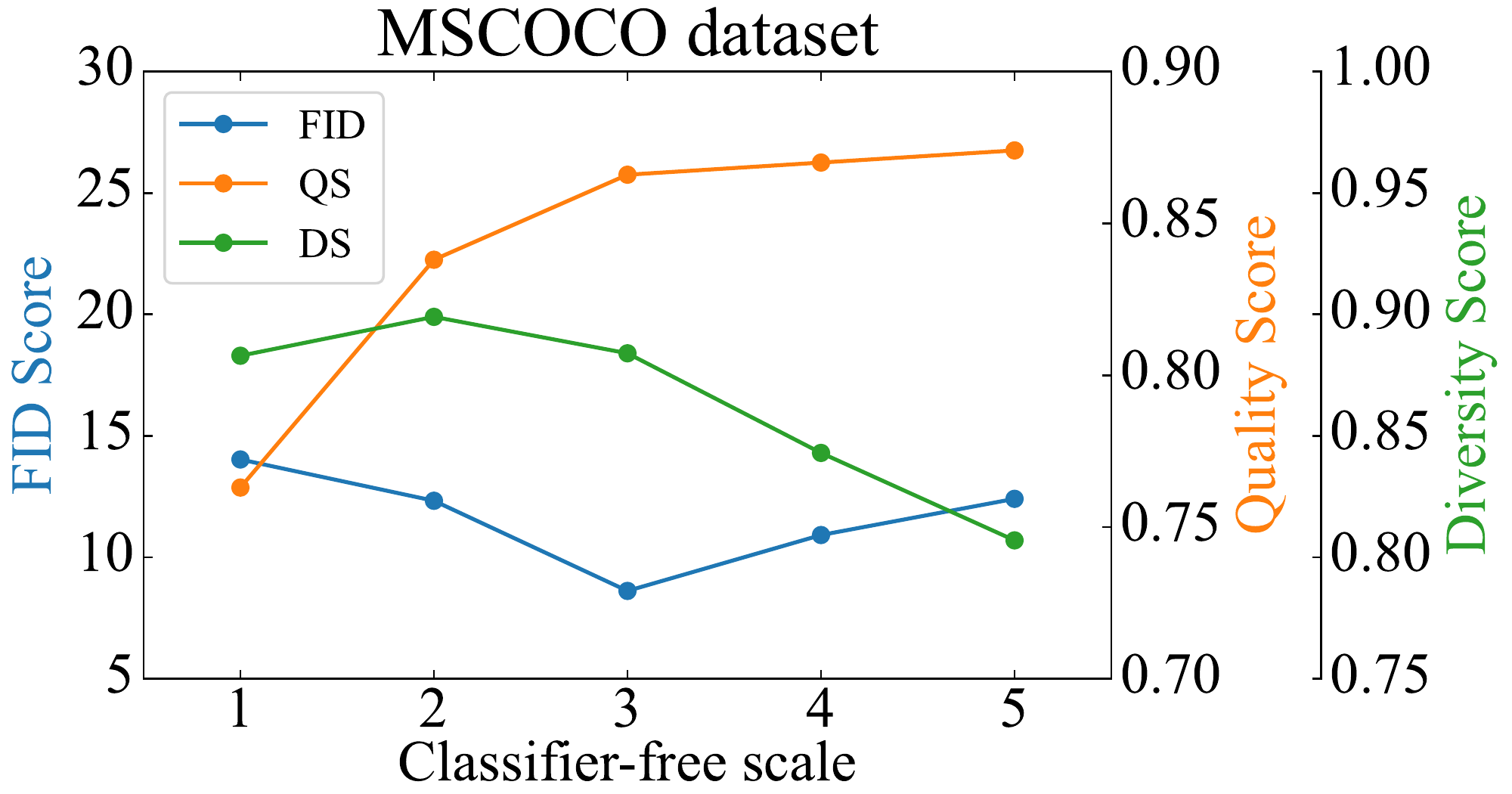}
\end{minipage}
\hfill
\begin{minipage}[c]{0.495\linewidth}
\includegraphics[width=\linewidth]{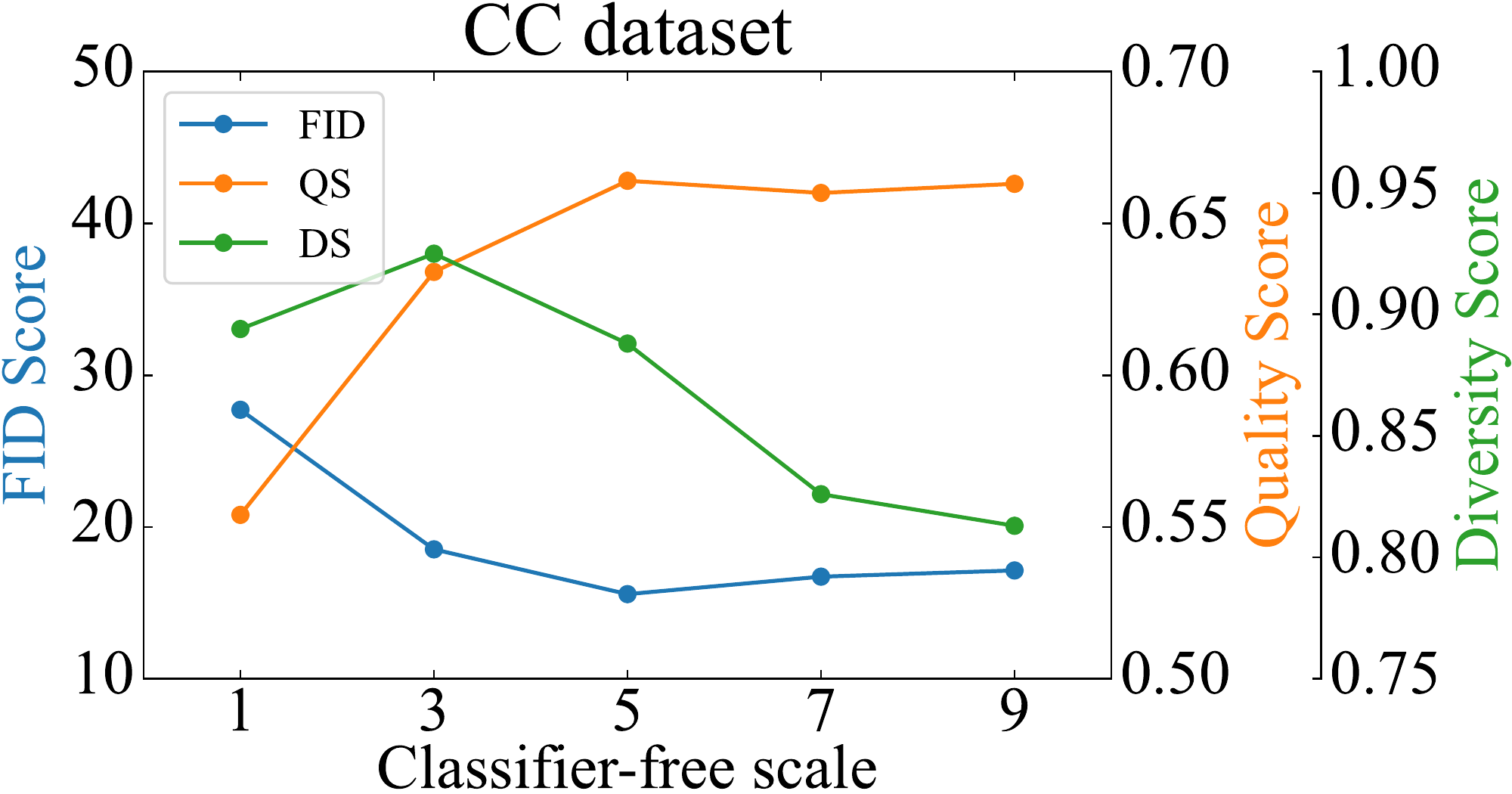}
\end{minipage}%
\vspace{-0.1cm}
\caption{Ablation study on classifier-free scale. Image quality rises as guidance scale increases, while a large scale may cause the loss of image diversity.}
\label{fig:cf}
\end{figure}

\begin{figure*}[t]
\centering
\includegraphics[width=1.0\columnwidth]{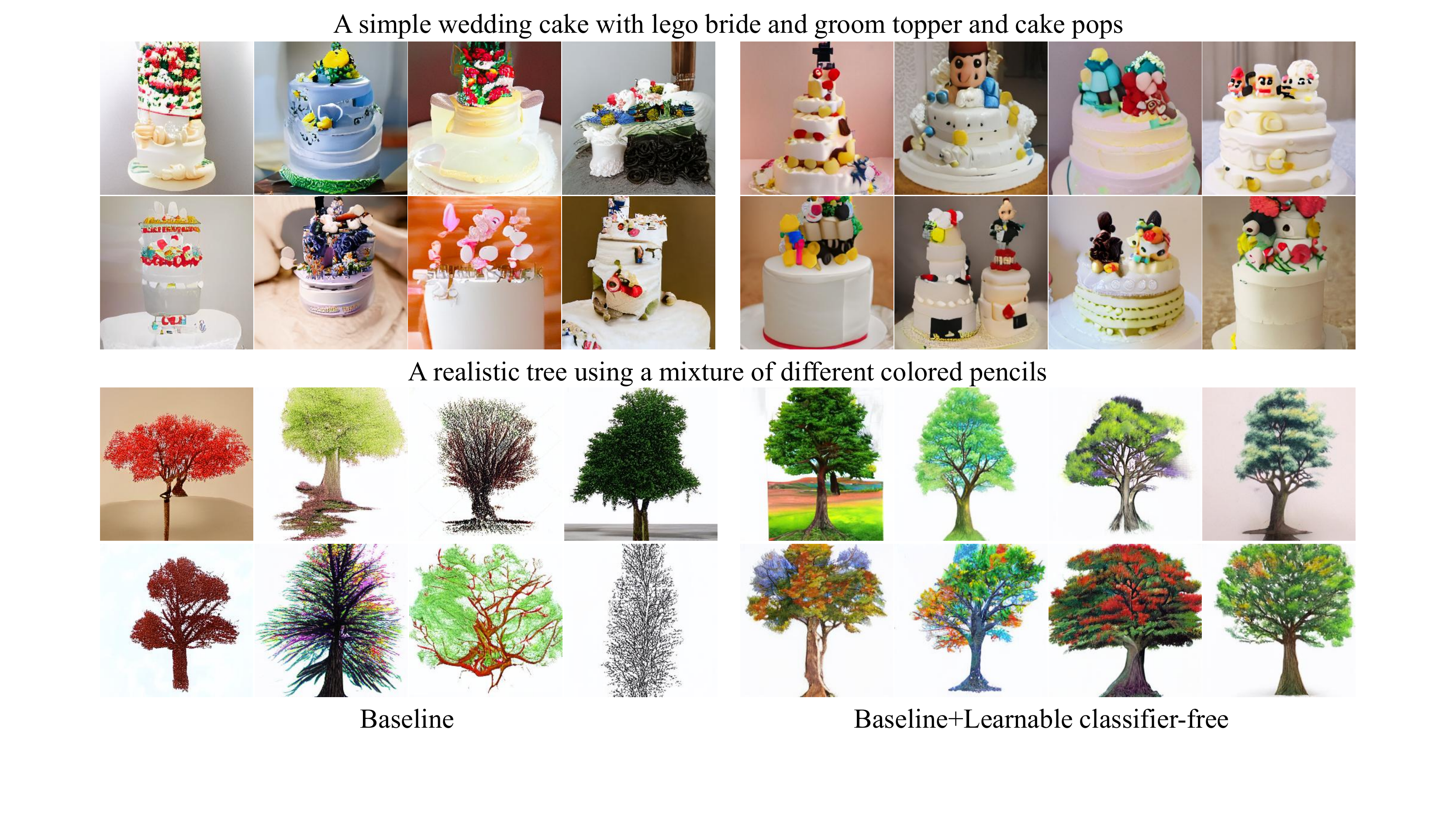}
\caption{Results of learnable classifier-free sampling on CC dataset.}
\label{fig:coco_learnable}
\end{figure*}

\subsection{Ablation Studies}
\paragraph{Discrete classifier-free guidance}
We investigate how the discrete classifier-free guidance could improve performance. We conduct the experiments on  MSCOCO and CC datasets. Specifically, we compare four different settings: 1) The original VQ-Diffusion model. 2) Without fine-tuning the models, we directly set the conditional input to null vector in the inference stage and apply the classifier-free sampling strategy. 3) We set $10\%$ of conditional input to null vector in the fine-tuning stage, and use the classifier-free guidance (Equation~\ref{eqn:cfsi}) to generate images. 4) Instead of setting conditional input to null vector, we set $10\%$ of them to a learnable vector to fit the unconditional image. 

We show the results in Table~\ref{table:cf}. We can find that classifier-free sampling could improve both the FID score and the Quality Score. Besides, the zero-shot classifier-free sampling strategy could improve performance without any training. By fine-tuning this model, the performance could be further improved. And using a learnable vector could achieve better performance than null vectors.

We also investigate how the guidance scale $s$ affects the results. As shown in Fig~\ref{fig:cf}, we conduct experiments on MSCOCO and CC datasets. We find that when $s$ increases, the Quality Score becomes better and the Diversity Score decreases. It demonstrates that the classifier-free sampling is a balance between quality and diversity. Besides, the FID score achieves the best performance when $s$ equals to 3 on MSCOCO, and 5 on CC dataset. Fig~\ref{fig:coco_learnable} provides a qualitative comparison of this posterior constraint. We can find by using the learnable classifier-free sampling strategy, the quality of generated images is improved significantly, and they are more correlated to the input text. Meanwhile, the diversity decreases.

\begin{table*}[h]
\centering
\renewcommand\arraystretch{1.2}
  \caption{Comparison of different classifier-free methods. Classifier-free outperforms baseline by a large margin and the learnable method further pushes performance.}
  \begin{tabular}{l|ccc|ccc}
    \hline
    \multirow{2}{*}{ } &
      \multicolumn{3}{c|}{MSCOCO } &   
      \multicolumn{3}{c}{CC} \\
    \cline{2-7}
    &  FID$\downarrow$ &  QS$\uparrow$ &  CLIP$\uparrow$  & FID$\downarrow$ & QS$\uparrow$ & CLIP$\uparrow$  \\
	\shline
\!\!\!\!  VQ-Diffusion     &      13.86       &   0.841 & 0.267 &   33.65  & 0.586 & 0.257 \\
\hline
\!\!\!\!  VQ-Diffusion + zero-shot CF      &    12.12    &  0.845 & 0.284 &  25.51 & 0.661 & 0.292 \\
\!\!\!\!  VQ-Diffusion + CF & 8.85  & 0.864 & 0.302 & 16.44 & 0.647 & 0.298 \\
\!\!\!\!  VQ-Diffusion + learnable CF & \textbf{8.62}  & \textbf{0.866} & \textbf{0.304} & \textbf{15.58} & \textbf{0.665} & \textbf{0.304} \\
	 \shline
  \end{tabular}
\label{table:cf}
\end{table*}

\paragraph{High-quality inference strategy}
Previous work~\cite{gu2021vector} proposed the fast sampling strategy which adopt inference steps fewer than training. We investigate the high-quality inference strategy in Table~\ref{table:slowinf} which use more inference steps than training to achieve better performance. We perform the experiment on the CUB-200 dataset and evaluate the generated images of 25,50,100,200 inference steps on five models with different training steps. We find by increasing the inference steps, the diffusion model could generate images with a better FID score. And the performance continues to get better as the inference steps increase.

\definecolor{graylight}{gray}{0.85}
\begin{table}[h]
\centering
\renewcommand\arraystretch{1.2}
\caption{FID score of high-quality inference strategy on CUB-200. The shaded part denotes fast inference strategy from ~\cite{gu2021vector}.}
\begin{tabular}{c|c|c|c|c|c|c}
\hline
\multirow{7}{*}{\rotatebox{90}{\!\!\!\!\!\!\!\!\! Inference steps}}&
\multicolumn{6}{c}{Training steps} \\
\shline
&  & 10&25&50&100&200\\
\cline{2-7}
  & 10 &32.35  & \cellcolor{graylight}27.62  & \cellcolor{graylight}23.47 & \cellcolor{graylight}19.84  &  \cellcolor{graylight}20.96 \\
\cline{2-7}
  & 25 & 26.89        &18.53  &  \cellcolor{graylight}15.25 & \cellcolor{graylight}14.03  &  \cellcolor{graylight}16.13 \\
\cline{2-7}
  & 	 50 & 22.70        & 16.93      & 13.82  & \cellcolor{graylight}12.45 & \cellcolor{graylight}13.67   \\
\cline{2-7}
  & 	 100 & 20.85      & 15.76      & 12.34        & 11.94 & \cellcolor{graylight}12.27 \\
\cline{2-7}
  & 	 200 & 19.99      & 15.55      & 12.20        & 11.87         & 11.80 \\
\shline
\end{tabular}
\label{table:slowinf}
\end{table}

\paragraph{Purity prior sampling}
We investigate the improvement of purity prior sampling. A token with higher purity demonstrates it has higher confidence to leave the $[\text{\tt{MASK}}]$ state. We conduct experiments on MSCOCO, CUB-200, CC and ITHQ-200M datasets. As shown in Table~\ref{table:prior}, we find that by adding the prior, all of these results are improved. Especially on larger datasets(CC and ITHQ-200M), the improvement is more significant. Meanwhile, this strategy requires neither training nor additional inference time. So it is an effective sampling strategy to improve the performance.

\begin{table}[h]
\centering
\renewcommand\arraystretch{1.2}
\caption{FID score of purity prior sampling strategy.}
\begin{tabular}{l|c|c|c|c}
    \hline
                  &  MSCOCO &  CUB-200 & CC & ITHQ-200M \\ 
	\shline
   \!\!\!\!  VQ-Diffusion     &    13.86   &   10.32  &   33.65  &   25.87    \\
   \hline
   \!\!\!\!  VQ-Diffusion + prior & \textbf{13.79} & \textbf{10.21} & \textbf{33.09} & \textbf{25.15} \\
	 \shline
\end{tabular}
\label{table:prior}
\end{table}

\subsection{Compare with state-of-the-art methods}
\begin{figure*}[t]
\centering
\includegraphics[width=1.0\columnwidth]{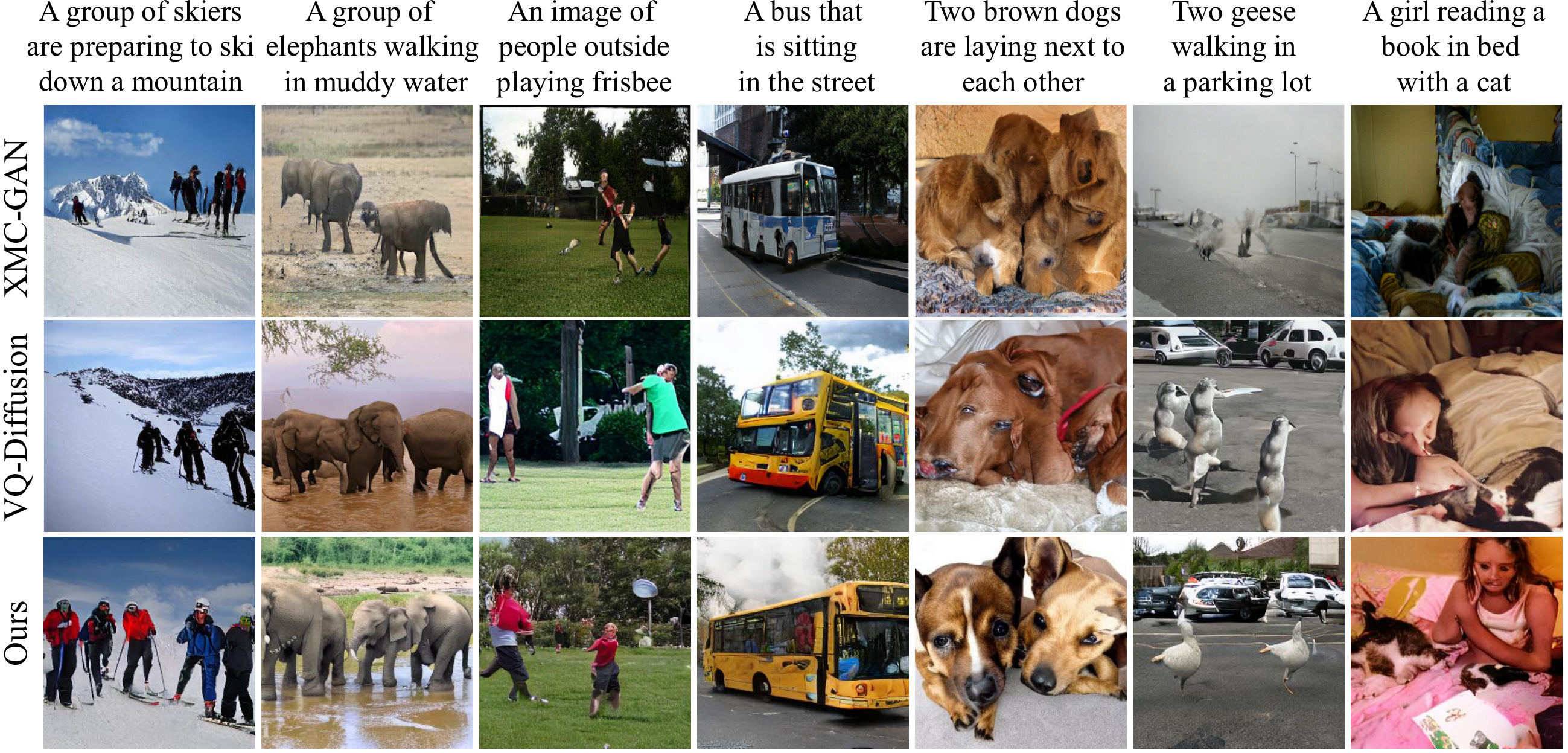}
\caption{Qualitative comparison with previous works on MSCOCO dataset.}
\label{fig:comparison}
\end{figure*}

\begin{figure*}[t]
\centering
\includegraphics[width=1.0\columnwidth]{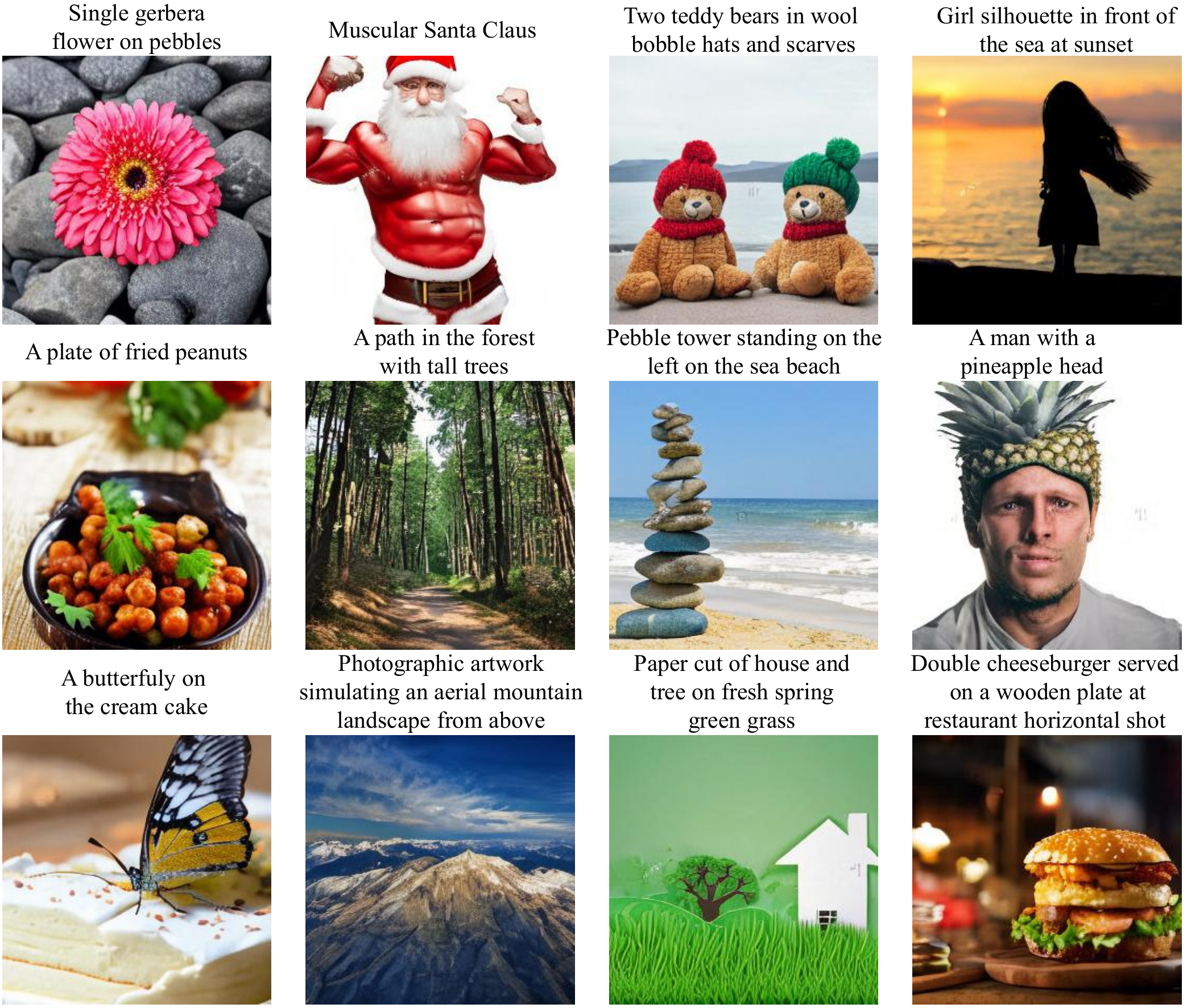}
\caption{In-the-wild text-to-image synthesis results of Improved VQ-Diffusion.}
\label{fig:vis}
\end{figure*}

We compare the proposed method with several state-of-the-art text-to-image methods, including DF-GAN~\cite{tao2020df}, XMC-GAN~\cite{zhang2021cross}, DALL-E~\cite{ramesh2021zero}, GLIDE~\cite{nichol2021glide}, and VQ-Diffusion~\cite{gu2021vector}, on MSCOCO and ITHQ-200M dataset. We evaluate FID score and show the results in Table~\ref{table:result} (a) and (c) respectively. Without any fine-tuning, we may leverage the zero-shot classifier-free sampling strategy and high-quality inference strategy to improve the performance of a well-trained VQ-Diffusion model, which is denoted as "Improved VQ-Diffusion*" in the table. Besides, by fine-tuning this model with the learnable classifier-free strategy, the performance is further improved.
We provide the visualized comparison with previous works on MSCOCO datasets in Fig~\ref{fig:comparison}, where our method could generate significantly better results. Besides, we provide the visualization results of the in-the-wild text-to-image synthesis results on VQ-Diffusion-L model in Fig~\ref{fig:vis}. Our method could generate very impressive results.

The proposed improved VQ-Diffusion is general, as it can also be applied to class conditional ImageNet generation tasks. We compare with BigGAN~\cite{brock2018large}, VQGAN~\cite{esser2021taming}, ImageBART~\cite{esser2021imagebart}, and VQ-Diffusion~\cite{gu2021vector}. The results are shown in Table~\ref{table:result} (b). Our Improved VQ-Diffusion achieves the best result among all compared methods.

\newcommand{\ft}[1]{{\fontsize{8.5pt}{8.5pt}\selectfont #1}}
\begin{table}[h]
     \caption{Comparison with previous methods on various datasets in terms of FID score. * indicates that we use only zero-shot classifier-free sampling and purity prior sampling.}
     \hspace{-0.2cm}
    \begin{subtable}[h]{0.28\textwidth}
        \centering
        \footnotesize
        \renewcommand\arraystretch{1.12}
        \begin{tabular}{l | c}
        \hline
         \ft{Methods} & FID$\downarrow$  \\
        \shline
        \ft{DFGAN~\cite{tao2020df}} & \ft{21.42} \\
        \ft{GLIDE~\cite{nichol2021glide}} & \ft{12.24} \\
        \ft{XMC-GAN~\cite{zhang2021cross}} & \ft{9.33} \\
        \ft{VQ-Diffusion~\cite{gu2021vector}} & \ft{13.86} \\
        \shline
        \ft{Improved VQ-Diffusion*} & \ft{11.89} \\
        \ft{Improved VQ-Diffusion} & \textbf{\ft{8.44}} \\
        \shline
        \end{tabular}
        \caption{MSCOCO}
    \end{subtable}
    \hspace{0.75cm}
    \begin{subtable}[h]{0.28\textwidth}
        \centering
        \footnotesize
        \renewcommand\arraystretch{1.12}
        \begin{tabular}{l | c}
        \hline
        \ft{Methods}  & \ft{FID$\downarrow$}  \\
        \shline
        \ft{ImageBART~\cite{esser2021imagebart}} & \ft{21.19} \\
        \ft{VQGAN~\cite{esser2021taming}} & \ft{15.78} \\
        \ft{BigGAN~\cite{brock2018large}} & \ft{7.53} \\
        \ft{VQ-Diffusion~\cite{gu2021vector}} & \ft{11.89} \\
        \shline
        \ft{Improved VQ-Diffusion*} & \ft{7.65} \\
        \ft{Improved VQ-Diffusion} & \textbf{\ft{4.83}} \\
        \shline
        \end{tabular}
        \caption{ImageNet}
    \end{subtable}
    \hspace{0.75cm}
    \begin{subtable}[h]{0.28\textwidth}
        \centering
        \footnotesize
        \renewcommand\arraystretch{1.12}
        \begin{tabular}{l | c}
        \hline
        \ft{Methods}  & \ft{FID$\downarrow$}  \\
        \shline
        \ft{VQ-Diffusion~\cite{gu2021vector}} & \ft{25.87} \\
        \shline
        \ft{Improved VQ-Diffusion*} & \ft{22.16} \\
        \ft{Improved VQ-Diffusion} & \textbf{\ft{19.06}} \\
        \shline
        \end{tabular}
        \caption{ITHQ-200M}
    \end{subtable}
     \label{table:result}
\end{table}

\section{Related Works}
\paragraph{Text-to-Image Synthesis}
In the past few years, Generative Adversarial Networks (GANs) have inspired many advances in image synthesis~\cite{reed2016generative,zhang2017stackgan,dash2017tac,gu2019mask,zhang2018photographic,gao2019perceptual,lao2019dual,zhang2021cross,ruan2021dae,el2019tell,cheng2020rifegan,souza2020efficient}. For text-to-image generation, most works could generate high fidelity images on single domain datasets, e.g., birds~\cite{wah2011caltech} and flowers~\cite{nilsback2008automated}. For more complex scenes with multiple objects, such as MSCOCO dataset~\cite{lin2014microsoft}, most GAN-based methods struggled to model these complicated distributions.

Other approaches are the Autoregressive Models (ARs). These models convert the complex distributions into multiple conditional distributions relying on previous outputs, thus have a stronger capability to model complex distributions. In recent works, DALL-E~\cite{ramesh2021zero} adopts a VQVAE to embed images to discrete tokens, and uses an AR model to fit the token distribution. It achieves impressive results on text-to-image generations. Other works like Cogview~\cite{ding2021cogview}, and M6~\cite{lin2021m6} have also achieved very promising results based on autoregressive models.

\paragraph{Denoising Diffusion Probabilistic Model}
Recently, an emerging generative model, the denoising diffusion probabilistic model (DDPM), has attracted great attention in the community. It has achieved strong results on image generation~\cite{ho2020denoising,nichol2021improved,ho2021cascaded,dhariwal2021diffusion,ashual2022knn,saharia2021palette}, video generation~\cite{ho2022video}, and audio generation~\cite{kong2020diffwave,popov2021grad}. It was first proposed in~\cite{ho2020denoising}, and the subsequent work~\cite{nichol2021improved} proposes a reparameterization to stabilize the training. ~\cite{nichol2021glide} proposed to use a classifier to guide the sampling process, which significantly improved the image quality. ~\cite{ho2021classifier} further discarded the classifier by introducing an additional fine-tune procedure. DALL-E 2~\cite{ramesh2022hierarchical} applied DDPM to massive data and achieved very impressive results on text-to-image synthesis tasks.

Meanwhile, other researchers investigate the discrete modeling approaches. D3PMs~\cite{austin2021structured} propose to use a transition matrix instead of Gaussian to add noise on discrete distributions. VQ-Diffusion~\cite{gu2021vector} uses a VQVAE to encode images into discrete tokens and leverages the discrete DDPM to model the discrete distribution. It achieved very strong performance on many image synthesis tasks. MaskGIT~\cite{chang2022maskgit} also predicts discrete tokens with the technique of masked pre-training. 

\section{Conclusion}
In this paper, we carefully studied VQ-Diffusion and identify it suffers from two main issues: the posterior issue and the joint distribution issue. To address these issues, we propose two techniques and improve the quality of the generated samples and their consistency with the input text by a large margin. Moreover, our strategies can even benefit VQ-Diffusion without fine-tuning the model. We demonstrate the superiority of our methods on various datasets. We hope our work opens the path for exploring VQ-Diffusion and facilitating future research.  

{\small
\bibliographystyle{ieee_fullname}
\bibliography{egbib}
}



\newpage

\appendix

\section{Proof of Lemma~\ref{lemma:mask}}

\lemmamask*

\emph{Proof}. For any position $i$ that satisfied $x_t^i = [\text{\tt{MASK}}]$, 

\begin{equation}
    \begin{split}
         &\ q(x_{t-1}^i = [\text{\tt{MASK}}] | x_t^i = [\text{\tt{MASK}}], x_0^i) \\
        =&\ q(x_{t-1}^i = K+1 | x_t^i = K+1, x_0^i) \\
        =&\ \frac{(v^{T}(x_t^i)\bm{Q}_{t}v(x_{t-1}^i))(v^{T}(x_{t-1}^i)\bar{\bm{Q}}_{t-1}v(x_0^i))}{v^{T}(x_t^i)\bar{\bm{Q}}_tv(x_0^i)}  \\
        =&\ \frac{(v^{T}(K+1)\bm{Q}_{t}v(K+1))(v^{T}(K+1)\bar{\bm{Q}}_{t-1}v(x_0^i))}{v^{T}(K+1)\bar{\bm{Q}}_tv(x_0^i)} \\
        =&\ \frac{v^{T}(K+1)\bar{\bm{Q}}_{t-1}v(x_0^i)}{v^{T}(K+1)\bar{\bm{Q}}_tv(x_0^i)}.
    \end{split}
\end{equation}

\noindent Since $x_0$ is the unnoised state, we know that $x_0^i \ne [\text{\tt{MASK}}]$. So we have

\begin{equation}
    \begin{split}
         &\ q(x_{t-1}^i = [\text{\tt{MASK}}] | x_t^i = [\text{\tt{MASK}}], x_0^i) \\
        =&\ \frac{v^{T}(K+1)\bar{\bm{Q}}_{t-1}v(x_0^i)}{v^{T}(K+1)\bar{\bm{Q}}_tv(x_0^i)} \\
        =&\ \frac{\bar{\gamma}_{t-1}}{\bar{\gamma_{t}}}.
    \end{split}
\end{equation}

\section{High-quality inference strategy Details}

\subsection{Fewer tokens sampling strategy}

Considering $\beta_t$ in the probability transition matrix $\bm{Q}_t$ is set to a tiny scale in implementations, we can ignore the replace situation. The detailed fewer tokens sampling strategy is shown in Algorithm~\ref{alg:fewer}. For each timestep $t$, we first denote all the locations with $[\text{\tt{MASK}}]$ as set $A_t=\{i|x_t^i = [\text{\tt{MASK}}]\}$. Then, we choose $\Delta_z$ items from $A_t$ by random sampling, we denote this set as $C_t$. Finally, we sample $\bm{x}_{0,t}$ from $p_\theta(\tilde{\bm{x}}_0|\bm{x}_t, \bm{y})$ and replace the token $\bm{x}_t^i$ with $\bm{x}_{0,t}^i$ for all locations $i\in C_t$. Sampling $\bm{x}_{0,t}$ from $p_\theta(\tilde{\bm{x}}_0|\bm{x}_t, \bm{y})$ has a similar effect as sampling from $p_\theta(\bm{x}_{t-1}|\bm{x}_t, \bm{y})$, but it could ensure to leave the $[\text{\tt{MASK}}]$ state.

\begin{algorithm}[h]
	\caption{Fewer tokens sampling strategy. Assume sampling $\Delta_z$ tokens each time, input text $s$.} \label{alg:fewer}
	\begin{algorithmic}[1]
		\State $t \gets T$, $\bm{y} \gets \text{BPE}(s)$
		\State $\bm{x}_t \gets \text{sample from } p(\bm{x}_T)$
		\While{$t > 0$}
		\State $A_t=\{i|x_t^i = [\text{\tt{MASK}}]\}$
		\State $C_t=\{c_1,c_2,\dots,c_{\Delta_z}\} \gets \text{random sample from } A_t$
		\State $\bm{x}_{0,t} \gets \text{sample from }p_\theta(\tilde{\bm{x}}_0|\bm{x}_t, \bm{y})$
		\For{$i=c_1,c_2,\dots,c_{\Delta_z}$}
		\State $x_t^i \gets x_{0,t}^i$
		\EndFor
		\State $t \gets \argmin_{t}||\frac{|A|-\Delta_z}{H\times W} - \bar{\gamma_t}||_2$
		\EndWhile
		\State \Return VQVAE-Decoder($\bm{x}_t$)
	\end{algorithmic}
\end{algorithm}

\subsection{Purity prior sampling strategy}

The detailed purity prior sampling strategy is shown in Algorithm~\ref{alg:prior}. It has similar implementation to the fewer tokens sampling but does not need to sacrifice the inference time, as we can set $\Delta_z$ according to normal inference timesteps.

It has two differences with Algorithm~\ref{alg:fewer}. 1) It calculates the purity for each location, and chooses locations with importance sampling rather than random sampling. 2) It adjusts the probability $\tilde{\bm{x}}_0$ with Equation~\ref{eqn:purity_adjust} since we find that larger kurtosis for locations with high purity helps improving quality. $r$ is a hyper-parameter and normally the range is $(0.5, 2)$ in our experiments.

\begin{equation}
\begin{split}
    \tilde{\bm{x}}_0 = \text{softmax}\left((1+purity_t\cdot r)\log\tilde{\bm{x}}_0\right)
\end{split}
\label{eqn:purity_adjust}
\end{equation}

\begin{algorithm}[h]
	\caption{Purity prior sampling strategy. Assume sampling $\Delta_z$ tokens each time, input text $s$, prior scale $r$.} \label{alg:prior}
	\begin{algorithmic}[1]
		\State $t \gets T$, $\bm{y} \gets \text{BPE}(s)$
		\State $\bm{x}_t \gets \text{sample from } p(\bm{x}_T)$ 
		\While{$t > 0$}
		\State $A_t=\{i|x_t^i = [\text{\tt{MASK}}]\}$
		\State $purity_t^i \gets \max_{j=1...K} p(\tilde{x}_0^i=j|x_t^i)$ \Comment{Eqn.~\ref{eqn:purity}}
		\State $C_t=\{c_1,c_2,\dots,c_{\Delta_z}\} \gets \text{importance sample from } A_t \text{ with } purity_t$
		\State $\tilde{\bm{x}}_0 \gets \text{softmax}\left((1+purity_t\cdot r)\log\tilde{\bm{x}}_0\right)$ \Comment{Eqn.~\ref{eqn:purity_adjust}}
		\State $\bm{x}_{0,t} \gets \text{sample from }p_\theta(\tilde{\bm{x}}_0|\bm{x}_t, \bm{y})$
		\For{$i=c_1,c_2,\dots,c_{\Delta_z}$}
		\State $x_t^i \gets x_{0,t}^i$
		\EndFor
		\State $t \gets \argmin_{t}||\frac{|A|-\Delta_z}{H\times W} - \bar{\gamma_t}||_2$
		\EndWhile
		\State \Return VQVAE-Decoder($\bm{x}_t$)
	\end{algorithmic}
\end{algorithm}

\section{Results}

\begin{figure*}[t]
\vspace{-1.0cm}
\centering
\includegraphics[width=1.0\columnwidth]{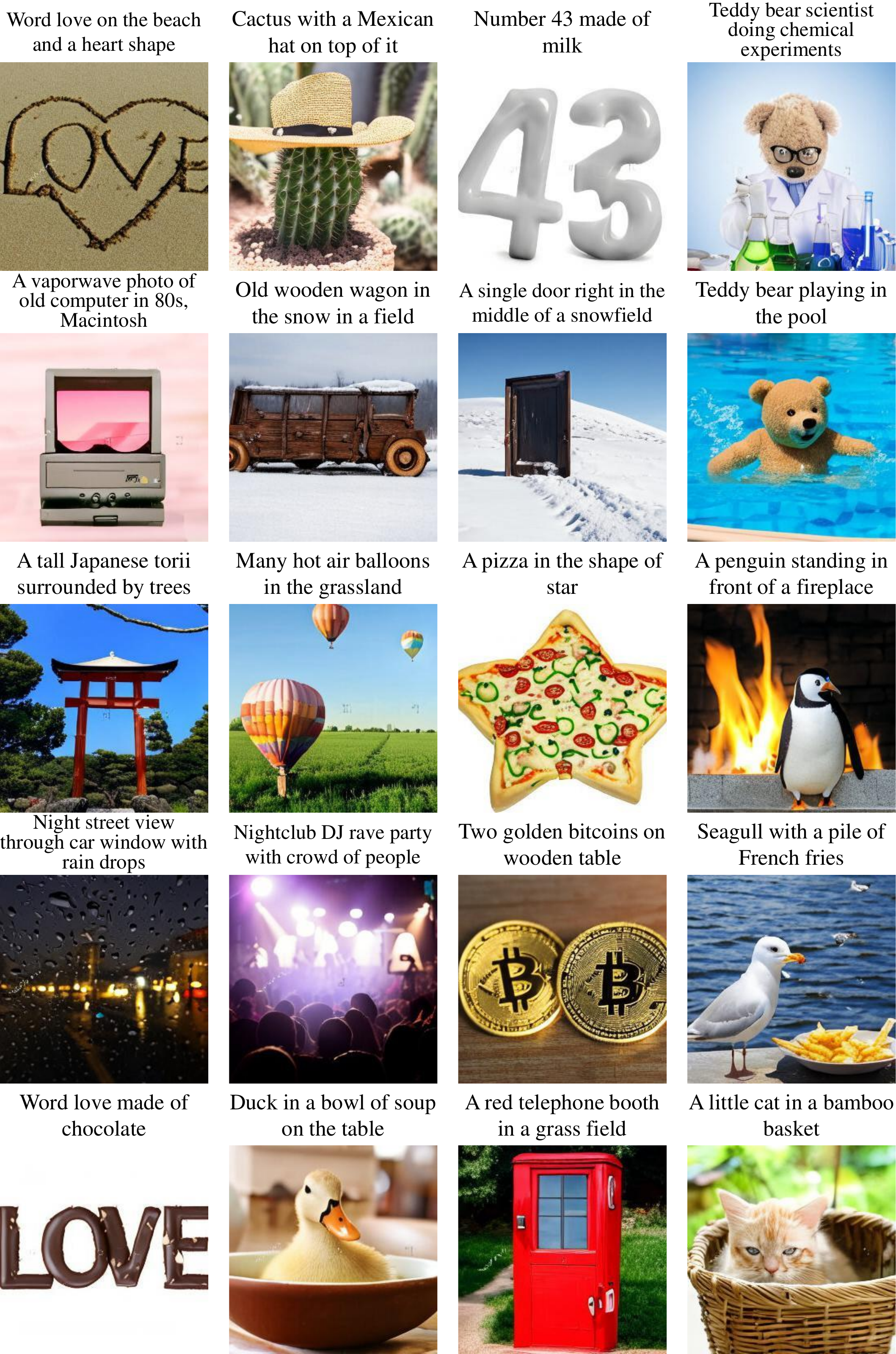}
\vspace{-0.5cm}
\caption{In-the-wild text-to-image synthesis with Improved VQ-Diffusion-L.}
\label{fig:in-the-wild-1}
\end{figure*}

\begin{figure*}[t]
\vspace{-1.0cm}
\centering
\includegraphics[width=1.0\columnwidth]{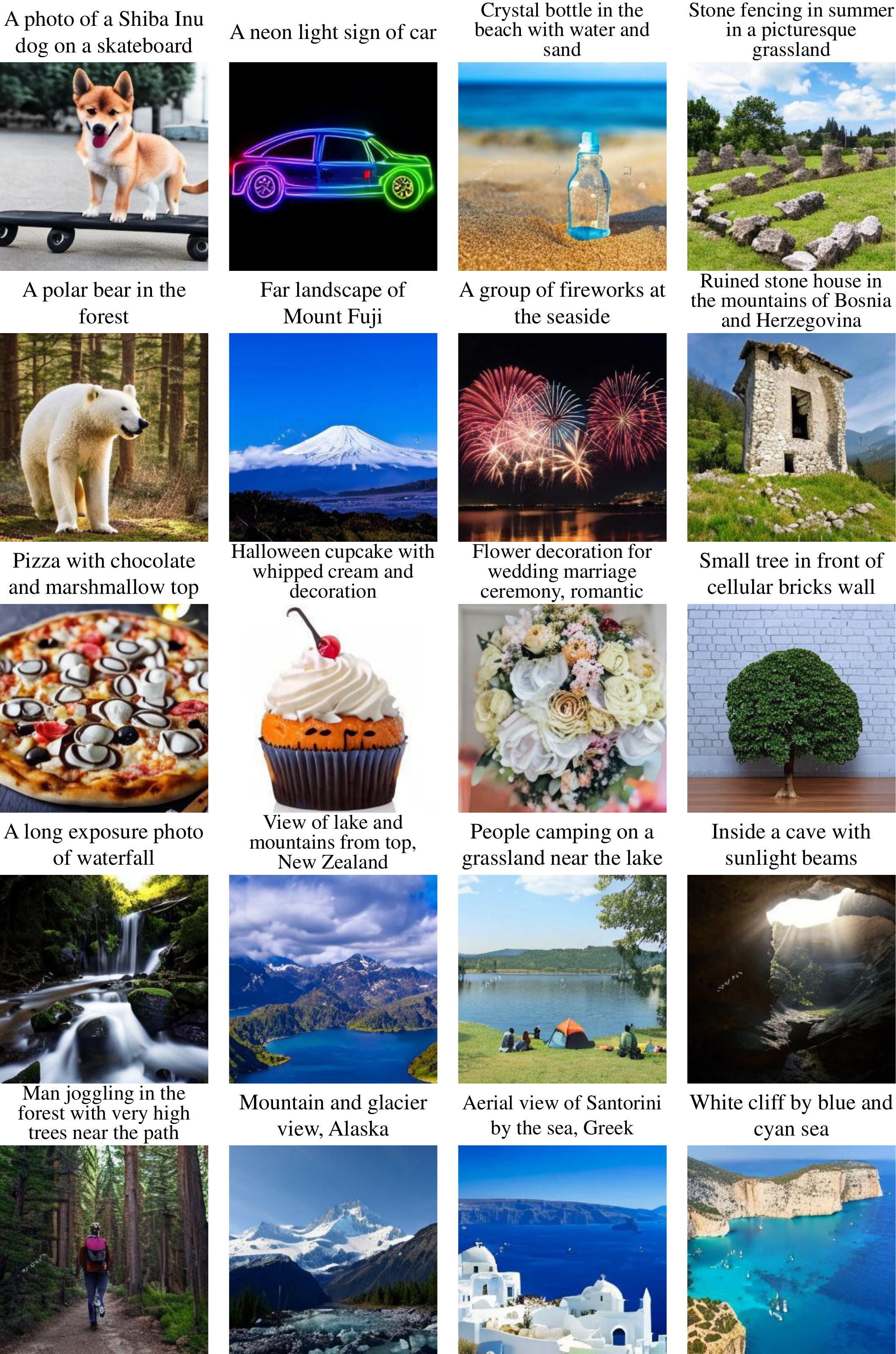}
\vspace{-0.5cm}
\caption{In-the-wild text-to-image synthesis with Improved VQ-Diffusion-L.}
\label{fig:in-the-wild-2}
\end{figure*}

\begin{figure*}[t]
\vspace{-1.0cm}
\centering
\includegraphics[width=1.0\columnwidth]{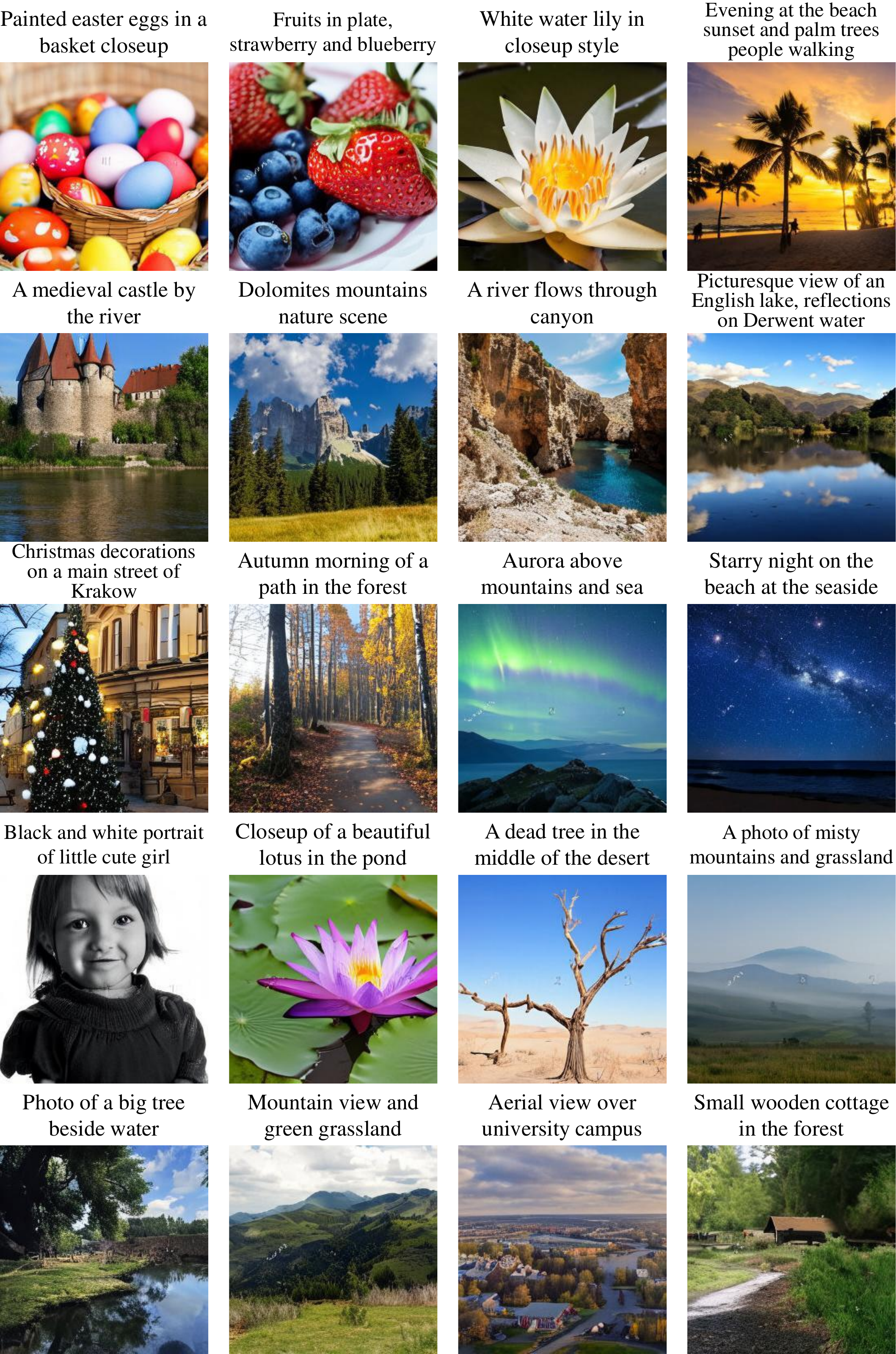}
\vspace{-0.5cm}
\caption{In-the-wild text-to-image synthesis with Improved VQ-Diffusion-L.}
\label{fig:in-the-wild-3}
\end{figure*}

In Figure~\ref{fig:in-the-wild-1}, Figure~\ref{fig:in-the-wild-2} and Figure~\ref{fig:in-the-wild-3}, we provide more visualization results of in-the-wild text-to-image synthesis on Improved VQ-Diffusion-L model.

\end{document}